\documentclass[final]{cvpr}

\usepackage{times}
\usepackage{epsfig}
\usepackage{graphicx}
\usepackage{amsmath}
\usepackage{amssymb}
\usepackage{xfrac}
\usepackage{bm}
\usepackage{siunitx}
\usepackage{multirow}
\usepackage{subcaption}
\usepackage[dvipsnames]{xcolor}

\usepackage[pagebackref=true,breaklinks=true,colorlinks,bookmarks=false]{hyperref}

\def\cvprPaperID{12} 
\def\confYear{CVPR 2021}

\begin{document}

\title{MVFuseNet: Improving End-to-End Object Detection and \\ Motion Forecasting through Multi-View Fusion of LiDAR Data}

\author{Ankit Laddha, Shivam Gautam, Stefan Palombo, Shreyash Pandey, Carlos Vallespi-Gonzalez\\
Aurora Innovation\\
{\tt\small aladdha,sgautam,spalombo,spandey,cvallespi@aurora.tech}
}

\maketitle



\maketitle

\begin{abstract}
In this work, we propose \textit{MVFuseNet}, a novel end-to-end method for joint object detection and motion forecasting from a temporal sequence of LiDAR data. Most existing methods operate in a single view by projecting data in either range view (RV) or bird's eye view (BEV). In contrast, we propose a method that effectively utilizes both RV and BEV for spatio-temporal feature learning as part of a temporal fusion network as well as for multi-scale feature learning in the backbone network. Further, we propose a novel sequential fusion approach that effectively utilizes multiple views in the temporal fusion network. We show the benefits of our multi-view approach for the tasks of detection and motion forecasting on two large-scale self-driving data sets, achieving state-of-the-art results. Furthermore, we show that MVFusenet scales well to large operating ranges while maintaining real-time performance.

\end{abstract}

\section{Introduction}
\vspace{-0.5em}
Object detection and motion forecasting are of paramount importance for autonomous driving. Object detection entails recognizing and localizing objects in the scene, whereas motion forecasting entails predicting the future trajectory of the detected objects. Traditionally, cascaded approaches treat detection and motion forecasting as two separate tasks, which enables classical methods for detection and motion forecasting to be used. However, these methods optimize for these tasks separately, suffering from cascading errors and missing the opportunity to share learned features for both tasks~\cite{spagnn}. To overcome these issues, multiple end-to-end methods have been proposed~\cite{intentnet,rvfusenet,multinet,liranet} for jointly solving both detection and motion forecasting. These methods have demonstrated excellent performance ~\cite{spagnn} while operating in real-time. Following the end-to-end paradigm, we propose a novel method for jointly detecting objects and predicting their future trajectories using time-series LiDAR data. 

\begin{figure}[!t]
    \centering
    \includegraphics[width=0.45\textwidth]{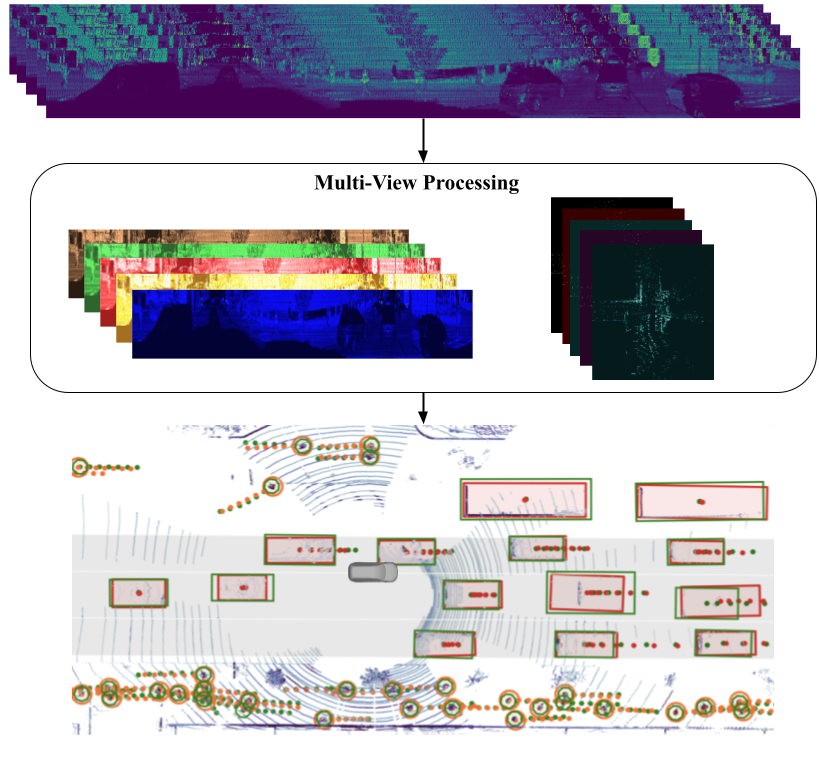}
\caption{The input to our method is a temporal sequence of 3D native range view images from LiDAR (top) and the output is object detections and motion predictions in the Cartesian bird's eye view (bottom). In contrast to previous single view methods, we propose to process the sequence in both views (middle).}
\label{fig:input_output}
\vspace{-1em}
\end{figure}

The input LiDAR data is natively captured in the  perspective range view (RV). However, since most planning algorithms operate in the Cartesian bird's eye view (BEV) space, the object detections and their forecasts need to also be in the same Cartesian space (see Figure ~\ref{fig:input_output}). Therefore, every method converts perspective RV information to a Cartesian BEV at some stage during its processing. Most existing methods lie on the extreme ends of the spectrum with respect to when they perform this conversion during their processing, and most use a single view entirely. On one hand, methods such as~\cite{rvfusenet,laserflow,lasernet} process LiDAR data exclusively in RV and only convert their final output to BEV during post processing. These methods are efficient for processing large spatial regions due to the compact size of the input image and offer state-of-the-art performance in the detection of small objects (e.g., pedestrians, bikes) and far away objects.  On the other hand, methods such as~\cite{intentnet,multinet,pointpillars} project the LiDAR data in the BEV first, with minimal or no pre-processing in RV, and perform most of the processing in BEV. The Cartesian BEV has the advantage of a strong prior due to range invariance of object shape and motion. This provides an edge to existing BEV methods on motion forecasting tasks; however, their scalability to operate in large areas remains a challenge. There has been some recent work on using multiple views for detection~\cite{multiview,fadadu2020multi}, but the space of models that can efficiently use multiple views for end-to-end detection and motion forecasting remains largely unexplored.

Given the complementary benefits of processing in both views, we posit that effectively combining both of them can lead to improved performance in both detection and forecasting. Therefore, in this work we propose \textit{MVFuseNet}, a novel end-to-end joint object detection and motion forecasting method which achieves state-of-the-art results on two large scale data sets and has real-time performance when processing a large spatial region. To accomplish this, we propose a novel sequential multi-view (MV) fusion network to aggregate a temporal sequence of LiDAR data for learning spatio-temporal features. We further propose a multi-view backbone network to process the spatio-temporal features for detection and forecasting. We demonstrate the effectiveness of multiple views over a single view on multiple data sets with different characteristics and LiDAR resolutions.

\begin{figure*}[ht]
    \centering
    \begin{subfigure}[b]{0.52\textwidth}
    \centering
    \includegraphics[width=\textwidth]{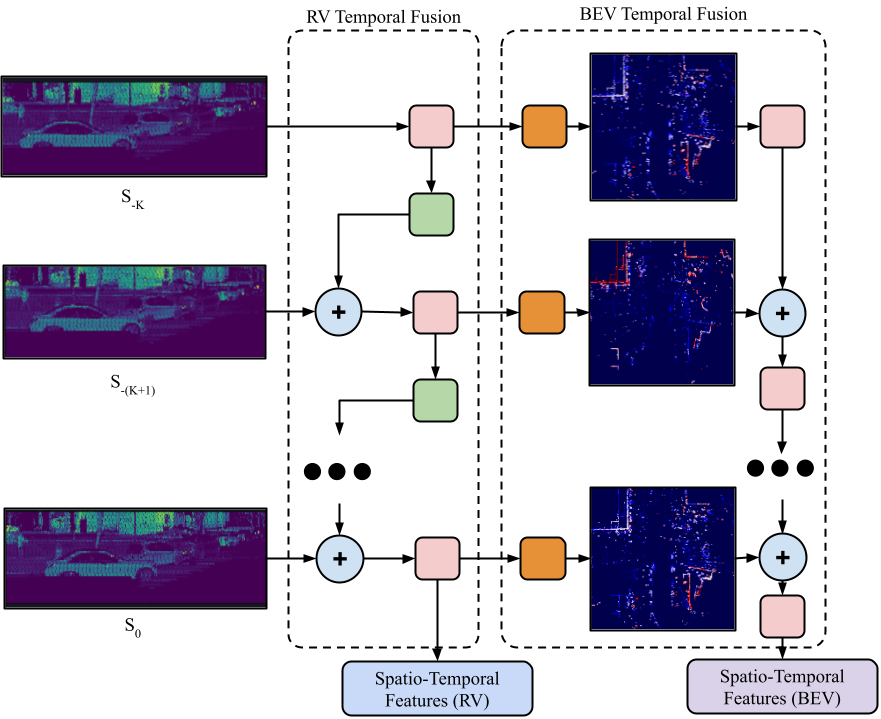}
    \caption{Multi-View Temporal Fusion Network}
    \label{fig:overview-tf}
    \end{subfigure}
    \begin{subfigure}[b]{0.42\textwidth}
    \centering
    \includegraphics[width=\textwidth]{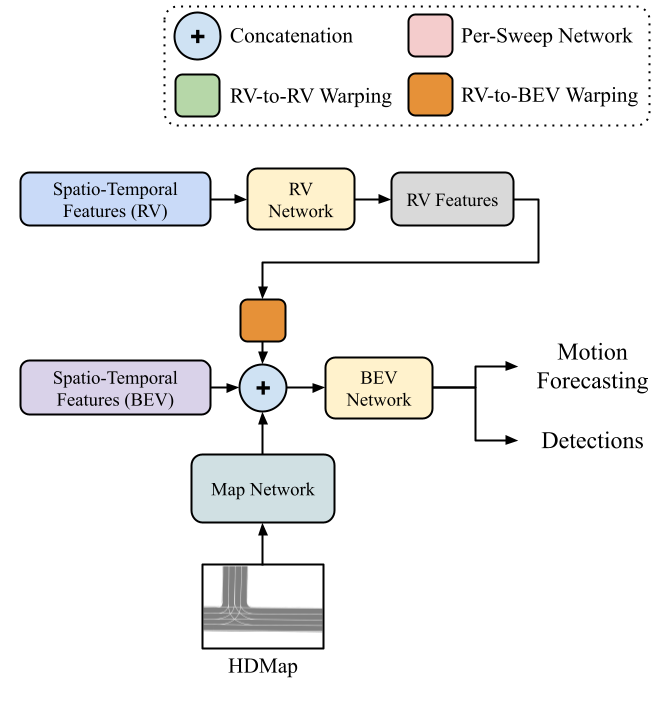}
    \caption{Multi-View Backbone Network}
    \label{fig:overview-bb}
    \end{subfigure}

\caption{\textbf{MVFuseNet Overview:} We propose a novel approach for (a) multi-view temporal fusion of LiDAR data in RV and BEV to learn spatio-temporal features. We sequentially aggregate sweeps by projecting the data from one sweep to the next sweep in the temporal sequence. (b) These multi-view spatio-temporal features are further processed by a multi-view backbone to combine them with map features and learn multi-scale features for final detection and motion forecasting.} 

\label{fig:overview}
\vspace{-1em}
\end{figure*}

\section{Related Work}
\vspace{-0.5em}
In this section, we first discuss the existing literature on LiDAR representation, and then look at various approaches for motion forecasting.

\subsection{LiDAR representation}
\vspace{-0.5em}
A spinning LiDAR captures data as a multi-channel image of range measurements. In the literature, these range measurements have been represented in various ways for processing: unstructured 3D point clouds \cite{pointrcnn, std}, 3D voxels \cite{zhouVoxelNetEndtoEndLearning2018, second}, a 2D BEV grid \cite{pointpillars, pixor,zhang2020polarnet}  and the native 2D RV grid \cite{lasernet,li2016vehicle,milioto2019rangenet++,kochanov2020kprnet}. The point cloud and voxel based methods are computationally expensive and do not scale well to highly dynamic and crowded outdoor scenes. In comparison, 2D BEV or RV grid based methods are efficient but only use a single view (either BEV or RV) for processing LiDAR data. Recent work has investigated the use of multiple views \cite{mv3d,zhou2020end,chen2020mvlidarnet,liong2020amvnet,liang2020rangercnn} and shown that the complementary benefits of both views improve performance. However, these methods use only one frame of LiDAR data and only solve perception tasks such as object detection and semantic segmentation. In contrast, we propose a method which aggregates data from multiple frames to jointly solve both detection and motion forecasting in an end-to-end method by utilizing both the BEV and RV. 

Recently, \cite{fadadu2020multi} proposed a multi-view approach for the joint task. In this method, the authors proposed fusing a single-frame RV projection with multiple frames of BEV projection, which improves object detection performance. This method, however, limits the temporal fusion of LiDAR data to BEV and only employs RV features of a single sweep, missing high resolution motion cues. In contrast, our proposed method performs spatio-temporal fusion of both BEV and RV features for multiple frames of LiDAR data. To the best of our knowledge, this is the first method that exploits multiple views for both temporal fusion and multi-scale backbone feature learning. We show that this leads to better detection and motion forecasting performance.

\subsection{Motion Forecasting}
\vspace{-0.5em}
Traditional learning-based motion forecasting approaches~\cite{dp,sociallstm,convolutionalsocialpooling,desire} use temporal sequences of detections~\cite{lasernet,pointpillars, zhou2020end,lasernet++} to learn spatio-temporal features for each object. Recent work in traditional motion forecasting has focused on improving the modeling of uncertainty through multi-modality~\cite{destination, zhao2020tnt, multipath, desire, covernet,rhinehart2019precog,mfp,cui2019multimodal} and interactions between actors and the scene~\cite{sociallstm,convolutionalsocialpooling,gao2020vectornet,yu2020spatio,gupta2018social,lanegcn}. In contrast, we look at the complementary problem of learning better spatio-temporal object features for forecasting using sensor data. Our proposed method can also benefit from many of the recent advances in the motion forecasting literature. However, to simplify the experimentation, we leave their incorporation to future work. These traditional methods are successful in capturing complex relationships and generating realistic longer-term forecasts, but they suffer from cascading error issues~\cite{spagnn} and lose out on the rich features learned from sensor data. These methods also work on a per-object basis, which makes them hard to scale to dense, urban environments. 

To address the issues with traditional forecasting approaches, the seminal work by \cite{faf} proposed to jointly solve both object detection and motion forecasting. \cite{intentnet} improved upon \cite{faf} by incorporating scene information using a semantic and geometric HDMap. Approaches such as \cite{multinet} and \cite{spagnn} build on top of \cite{intentnet} by adding an object-centric sub-network to refine future trajectories. These methods show that recent work on multi-modal predictions and the use of interaction graphs to model complex relationships can be easily extended to the framework of joint object detection and motion forecasting. \cite{liranet} and \cite{fadadu2020multi} are recent multi-sensor methods that build on top of \cite{multinet} by using radar and camera inputs respectively. These methods, by virtue of operating in BEV, lose out on high-resolution point information and are often limited by range of operation. RV based methods such as \cite{laserflow} and \cite{rvfusenet} overcome the limitation on operating range but are outperformed in the motion forecasting task by recent BEV based methods. In this work, we improve the joint framework by including multi-view representation in multiple parts of the network and achieve state-of-the-art performance on both object detection and motion forecasting while scaling to large areas of operation in real-time.

\section{MV Detection and Motion Forecasting}
\vspace{-0.5em}
Figure~\ref{fig:overview} shows an overview of our proposed approach. Our main contribution is an end-to-end object detection and motion forecasting method that processes the time-series LiDAR data in both range view and bird's eye view. We first describe prerequisite information related to the input and view-projections in Section~\ref{sec:input}. We then discuss our contribution of using multiple views for temporal fusion of a sequence of LiDAR data in Section~\ref{sec:temporal_fusion}. In Section~\ref{sec:backbone}, we discuss our contribution of a multi-view backbone network to extract per-cell features. Finally, we present our method for joint detection and motion forecasting using the per-cell features in section~\ref{sec:output}, followed by the loss functions used to train the model in Section~\ref{sec:loss}.

\subsection{Preliminaries}
\label{sec:input}
\vspace{-0.5em}
\textbf{Input:} Let us assume that we are given a time-series of $K+1$ sweeps, where each sweep contains all the LiDAR points from a full 360$^{\circ}$ rotation of a LiDAR sensor. This time series LiDAR data can be denoted by $\{\mathcal{S}_{k}\}_{k=-K}^{0}$, where $k=0$ is the most recent sweep and $-K \le k \le 0$ are the past sweeps. We term the most recent sweep as the \textit{reference sweep}. Each LiDAR sweep contains $N_{k}$ range measurements, which can be transformed into a set of 3D points, $\mathcal{S}_{k}= \{\bm{p}^{i}_{k} \}_{i = 1}^{N_k}$, using the pose (viewpoint) of the sensor $\mathcal{P}_{k}$ at the end of sweep capture. We assume that pose for each sweep is provided by an onboard localization system. Therefore, we can calculate the transformation of points from one viewpoint to another. We denote the $k$-th sweep transformed into the $n$-th sweep's coordinate frame as, $\mathcal{S}_{k,n} = \{\bm{p}^{i}_{k,n}\}_{i = 1}^{N_k}$, where each point $\bm{p}^{i}_{k,n}$ is represented by its 3D coordinates, $[x^{i}_{k,n}, y^{i}_{k,n}, z^{i}_{k,n}]^T$. In spherical coordinates the same point $\bm{p}^{i}_{k,n}$ can be represented using the radial distance $r^{i}_{k,n}$, azimuth $\theta^{i}_{k,n}$ and elevation $\phi^{i}_{k,n}$. Note that $\bm{p}^{i}_{k,n}$ represents the same LiDAR return as $\bm{p}^{i}_{k}$, only transformed into a different frame. 

\textbf{Projections:} For each point $\bm{p}^{i}_{k}$ captured at pose $\mathcal{P}_{k}$, the range view projection at pose $\mathcal{P}_{n}$ is defined by discretizing the azimuth and elevation angles of $\bm{p}^{i}_{k,n}$. Similarly, the bird's eye projection at pose $\mathcal{P}_{n}$ is the $x$ and $y$ coordinates of $\bm{p}^{i}_{k,n}$.

\textbf{Per-Point Features:} For each point $\bm{p}^{i}_{k}$ in $\mathcal{S}_{k}$, we define a set of associated features as concatenation of its coordinates in original viewpoint, $[x^{i}_{k,k}, y^{i}_{k,k}, z^{i}_{k,k}]^T$, coordinates in most recent viewpoint, $[x^{i}_{k,0}, y^{i}_{k,0}, z^{i}_{k,0}]^T$ and  the remission or intensity $e^{i}_{k}$ of the LiDAR return.


\subsection{Multi-View Temporal Fusion Network}
\vspace{-0.5em}
\label{sec:temporal_fusion}
The goal of the temporal fusion sub-network is to aggregate a time-series of LiDAR data in order to learn spatio-temporal features. The most straightforward approach, as employed by many previous works \cite{intentnet, multinet, laserflow}, is the \textit{one-shot} approach where all the data is accumulated in a single frame. All  points are first transformed into the frame defined by the reference pose and then the aggregation is done by projecting them in either BEV or RV. For multiple views this can be trivially extended by projecting the points in both BEV and RV for aggregation. However, directly projecting all the past LiDAR data into the RV of the most recent sweep leads to significant performance degradation due to heavy data loss in the projection step \cite{rvfusenet}. Therefore, instead of previous approaches that focus on \textit{one-shot} projection, we propose a novel sequential multi-view fusion approach to effectively aggregate the temporal LiDAR data. 

\begin{figure}[!t]
    \centering
    \includegraphics[width=0.45\textwidth]{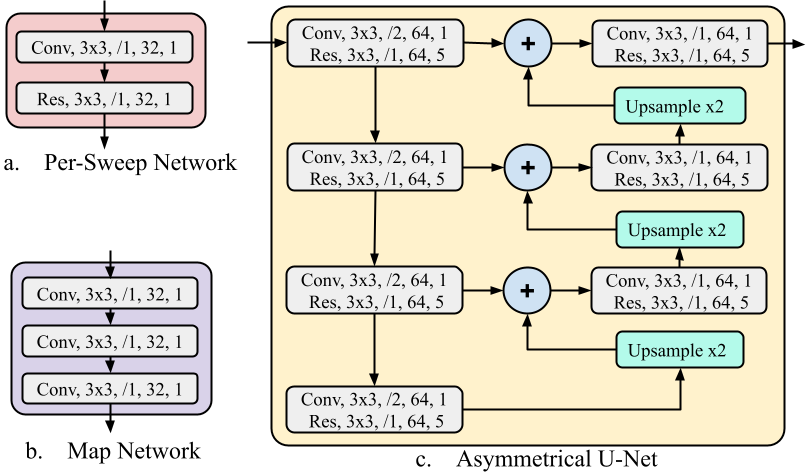}
\caption{\textbf{Network Components:} (a) We use the depicted per-sweep network to process each sweep during temporal fusion in both views. Note that no weights are shared across time and views during temporal fusion. (b) The HDMap is processed with the depicted network to learn local map-only features which are combined with the LiDAR features. (c) The asymmetric U-Net network is used to extract and combine multi-scale features in BEV. In RV, only the width dimension is down-sampled and the first convolutional layer is not strided. Each layer in the networks is represented as $B, k \times k,/s,C,N$, where $B$ is the block name, $k$ is the kernel size, $s$ is the stride, $C$ is the number of channels and $N$ is the number of repetitions of the block. \textit{Conv} denotes a convolutional layer followed by batch normalization and ReLU. \textit{Res} denotes a residual block as defined in~\cite{resnet}. Finally, we up-sample using bi-linear interpolation.}

\label{fig:arch}
\vspace{-1em}
\end{figure}

Figure~\ref{fig:overview-tf} shows our proposed fusion approach. We assume that the input is a time-series of multi-channel RV images in their original capture pose. These images contain the per point features, $\bm{f}^{i}_{k}$, as defined in Section~\ref{sec:input}. We sequentially fuse the LiDAR sweeps from one time-step to the next in both views. At each time-step we warp the previous time-step's RV features to the current time-step's frame (green box), and then use a sub-network (see Figure~\ref{fig:arch}\textcolor{red}{a}) to learn spatio-temporal features for each cell in RV (pink box). These learned features are then projected into the BEV (orange box) and concatenated with the BEV features from the previous time-step. Similarly to RV, a sub-network is then used to learn spatio-temporal features for each cell in BEV. The feature learning networks (pink box) in each view and time-step are independent and no weights are shared across time or view. It is important to note that unlike previous methods that project raw point-features to the BEV, our method projects learned RV features to be used in the BEV. We further discuss the methods used to warp features from one RV to another and for projecting the RV features to BEV.

\textbf{RV-to-RV Feature Warping:} Let us assume that we would like to warp the RV feature map $\mathcal{R}_{k,k}$ of $k$th sweep to the RV feature map $\mathcal{R}_{k,n}$ at viewpoint on $n$th sweep. We assume that the point $\bm{p}^{i}_{k}$ is projected to location $l^{i}_{k,k}$ in $\mathcal{R}_{k,k}$ and $l^{i}_{k,n}$ in $\mathcal{R}_{k,n}$. Therefore, we define the feature warping by copying the features from one RV to another such that $\mathcal{R}_{k,n}(l^{i}_{k,n}) = \mathcal{R}_{k,k}(l^{i}_{k,k})$. Similar to~\cite{rvfusenet, lasernet}, if more than one point project into the same cell location $l^{i}_{k,n}$, we pick the closest point for feature rendering.  

\textbf{RV-to-BEV Feature Warping:} Let us assume that we would like to warp the RV feature map $\mathcal{R}_{k,k}$ of $k$th sweep to the BEV feature map $\mathcal{B}_{k,0}$. We also assume that a point $\bm{p}^{i}_{k}$ in $\mathcal{S}_{k}$ can be projected in $\mathcal{R}_{k,k}$ to extract a learned feature $g^{i}_{k}$. We calculate the features of cell $l^{i}_{k,0}$ in $\mathcal{B}_{k,0}$ by aggregating the features of all the points $\mathcal{A}^{l}_{k} = \{ p^{i}_{k}, i=1,...,M\} $ that are projected into that cell location. Similarly to \cite{pointpillars, zhou2020end}, for each point in a cell, we calculate its feature vector $h^{i}_{k}$ by concatenating the difference between the coordinates of the point and the cell $\Delta c = [x^{i}_{k} - l^{i}_{x, k,0}, y^{i}_{k} - l^{i}_{y, k,0}]$, and the RV features of the point $g^{i}_{k}$. For aggregating the features of all the points in the cell we use:
\begin{equation}
    \mathcal{B}_{k,0}(l^{i}_{k,0}) =  \frac{1}{M}\sum_{i=0}^{M}\text{MLP}(h^{i}_k),
\end{equation}

where $\text{MLP}$ is a linear layer followed by batch normalization and ReLU. 

\subsection{Multi-View Backbone Network}
\label{sec:backbone}
\vspace{-0.5em}
The goal of the backbone is to process the spatio-temporal features and combine them with map features to learn per-cell features that can be used for object detection and motion forecasting. As shown in Figure~\ref{fig:overview-bb}, our backbone network processes the spatio-temporal features in both views. We first take the spatio-temporal features in RV as input and learn multi-scale RV features by extracting and combining features using an asymmetrical U-Net (see Figure~\ref{fig:arch}\textcolor{red}{c}). These RV features are then projected to BEV using the same technique as in Section~\ref{sec:temporal_fusion} and concatenated with learned map features and the spatio-temporal BEV features (see Figure~\ref{fig:overview-bb}). We rasterize the map features in BEV~\cite{intentnet,multinet} and learn high level features using a convolutional neural network (see Figure~\ref{fig:arch}\textcolor{red}{b}). Similar to RV, this multi-view, multi-sensor feature volume is further processed by another asymmetrical U-Net to learn multi-scale features in BEV (see Figure~\ref{fig:arch}\textcolor{red}{c}). 

\subsection{Output Prediction}
\label{sec:output}
\vspace{-0.5em}
Given the per cell features from the backbone network, our goal is to detect objects observed in the current sweep $\mathcal{S}_{0}$ and predict their trajectory. We use a dense, single-stage convolutional header for detecting
objects using the per-cell features. Similarly to ~\cite{multinet, centerdet}, we first predict if a cell contains the center of an object for some class. For each center cell, we then predict an associated bounding box and use non-maximum suppression to remove duplicates.  For motion forecasting of large objects such as vehicles, we extract a rotated region of interest (RROI)~\cite{spagnn,multinet} of $60\times60$m centered at the object to learn actor-centric features which are then used to predict the trajectory. However, for smaller objects such as pedestrians and bicycles, we use the features of the center cell to predict the trajectory since we empirically found that this leads to better results. 

\subsection{End-to-End Training}
\vspace{-0.5em}
\label{sec:loss}
Similarly to~\cite{multinet,liranet}. we train the proposed method end-to-end using a multi-task loss incorporating both detection and trajectory loss:  $\mathcal{L}_{total} = \mathcal{L}_{det} + \mathcal{L}_{traj}$.

\textbf{Detection Loss ($\mathcal{L}_{det}$)} is a multi-task loss defined as a weighted sum of classification and regression loss: $\mathcal{L}_{det} = \mathcal{L}^{cls}_{det} + \lambda \mathcal{L}^{reg}_{det}$. We use focal loss~\cite{retinanet} for classifying if a BEV cell is at the center of an object class. For each center cell, we use smooth L1 loss to learn parameters of the object bounding box relative to that cell. We parameterize each box $i$ by it's center $(x_{i}, y_{i})$, orientation ($\theta_{i}$) and size ($w_{i}, h_{i}$). The orientation is further parameterized as ($\cos(\theta_{i}), \sin(\theta_{i})$).

\textbf{Trajectory Loss ($\mathcal{L}_{traj}$)} is defined as an average of per future time-step loss: $\mathcal{L}_{traj} = \sfrac{1}{T}\sum_{t=1}^{T}\mathcal{L}^{KL}_{t}$~\cite{multinet}. We consider each waypoint at time $t$ of a trajectory $j$ to be a 2D Laplace distribution parameterized by its position ($x_{j}^{t}, y_{j}^{t}$) and scale ($\sigma_{j,x}^{t}, \sigma_{j,y}^{t}$). We use the KL divergence~\cite{lasernet-kl} between the ground truth and predicted distribution as loss $\mathcal{L}^{KL}_{t}$ to learn the per waypoint distribution. 

\begin{table*}[!ht]
    \centering
    \caption{\textbf{nuScenes}: Comparison of proposed \textit{MVFuseNet}, with existing end-to-end methods. The reported $L_2$ is at $3$s.}
    \vspace{-0.5em}
    \scalebox{0.85}{
    \begin{tabular}{c|c|c|c|c|c|c}
        \hline
        \multirow{2}{*}{Method} & \multicolumn{2}{c|}{Vehicle} & \multicolumn{2}{c|}{Pedestrian} & \multicolumn{2}{c}{Bikes} \\ \cline{2-7}
        & \multicolumn{1}{c|} {AP (\%) $\uparrow$} & \multicolumn{1}{c|}{$L_2$ (cm) $\downarrow$}  & \multicolumn{1}{c|} {AP (\%) $\uparrow$} & \multicolumn{1}{c|}{$L_2$ (cm) $\downarrow$}  & \multicolumn{1}{c|} {AP (\%) $\uparrow$} & \multicolumn{1}{c}{$L_2$ (cm) $\downarrow$} \\
        \hline
        SpAGNN~\cite{spagnn} & - & 145 & - & - & - & - \\
        \hline
        Laserflow~\cite{laserflow} & 56.1 & 143 & - & - & - & - \\
        RVFuseNet~\cite{rvfusenet} & 59.9 & 120 & - & - & - & - \\
        \hline
        LiRANet~\cite{liranet} & 63.7 & 102 & - & - & - & - \\
        \hline
        IntentNet~\cite{intentnet} & 60.3 & 118 & 63.4 & 84 & 31.8 & 173 \\
        MultiXNet~\cite{multinet} & 60.6  & 105 & 66.1 & 80 & 32.6 & 203\\
        \hline
        L-MV~\cite{fadadu2020multi} & 61.1  & 107 & 71.0 & 82 & 38.2 & 187\\
        \hline
        LC-MV~\cite{fadadu2020multi} & 62.9 & 107 & 71.4 & 80 & 39.8 & 179\\
        \hline
        MVFuseNet (Ours) & \textbf{67.8} & \textbf{99} & \textbf{76.4} & \textbf{75} & \textbf{44.5} & \textbf{138}\\
        \hline
    \end{tabular}
    }
    \label{tab:nuscenes}
    \vspace{-0.5em}
\end{table*}

\section{Experiments}
\vspace{-0.5em}
\subsection{Data set and Metrics}
\vspace{-0.5em}
We report results on two autonomous driving data sets, with different LiDAR resolutions and characteristics, to show the efficacy of our proposed approach. 
In particular, we use the publicly available nuScenes~\cite{nuscenes} data set, and a much larger internal data set. 
The nuScenes data set consists of $1$k snippets. 
It has a low resolution LiDAR which generates $\sim30$k points per sweep and a square region of interest (ROI) of length $100$m, centered on the self-driving vehicle (SDV). On the other hand, our internal data set consists of  $17$k snippets. It has a higher resolution LiDAR which generates $\sim130$k points per sweeep and uses a ROI of a square of $200$m length. On both data sets, we report results on three major classes of traffic participants: vehicles, pedestrians and bikes. 

Following previous works~\cite{spagnn,multinet,rvfusenet}, we use average precision (AP) with intersection over union (IoU) based association between ground truth and the detected object. Furthermore, we use $L_{2}$ displacement error at multiple time horizons to evaluate motion forecasting. We compute $L_{2}$ as the Euclidean distance between the center of the predicted true positive box and the associated ground truth box. Note that the official nuScenes leaderboard evaluates the task of detection and state estimation, whereas in this work we solve the joint task of detection and motion forecasting. Therefore, we use the same metrics as used in previous work~\cite{spagnn,intentnet} instead of the official leaderboard metrics.

\subsection{Implementation Details}
\vspace{-0.5em}
We use the PyTorch~\cite{paszke2019pytorch} library for implementing the proposed approach. On nuScenes, the input RV is chosen to be of size $32 \times 1024$ based on the LiDAR characteristics. Furthermore, the input BEV feature map is chosen to be $400 \times 400$ and the backbone output is chosen to be $200 \times 200$, to balance runtime and resolution. This results in an input resolution of $25$cm and an output resolution of $50$cm. On our internal data set, the input RV is $64 \times 2048$ and both the input BEV and output BEV feature map are of the size $400 \times 400$. Due to the large ROI, this results in a resolution of $50$cm at both input and output. For both data sets, we use the LiDAR data from the past $0.5$ seconds as input and predict the trajectory for $3$ seconds into the future, sampled at $10$Hz. Since nuScenes is much smaller than our internal data set, we use data augmentation during training. Specifically, we generate labels at non-key frames by linearly interpolating the labels at adjacent key frames. We further randomly augment each frame by applying translation ($\pm1$m for the $x$- and $y$-axes and $\pm0.2$m for $z$ axis) and rotation (between $\pm45^{\circ}$ along the $z$-axis) to both the point clouds and labels.

We train with a batch size of $64$ distributed over $32$ GPUs. We first pre-train the network without rotated ROI for $20$ epochs and then warm start the model with the pre-trained weights and train for $6$ more epochs. We train the network using a cosine learning rate schedule with a starting rate of $\num{1e-3}$ and an end rate of $\num{2e-5}$. We set the gamma in focal loss to $2$ and the loss weight $\lambda$ in the detection loss to $0.2$.

\subsection{Comparison to the State-of-the-Art}
\vspace{-0.5em}
In this section, we  compare our method to existing end-to-end methods using the evaluation setting of~\cite{multinet,fadadu2020multi}. As shown in Table~\ref{tab:nuscenes}, our novel multi-view method significantly outperforms all other methods, on both detection and forecasting tasks for all evaluated classes.

We see significant improvements on both detection and motion forecasting when we compare our method to the best RV-based method \textit{RVFuseNet}~\cite{rvfusenet}, and the state-of-the-art BEV-based method \textit{MultiXNet}~\cite{multinet}.  Notably, our method shows a $\sim15\%$ improvement on pedestrian detection, a $\sim40\%$ improvement on bike detection, and a $\sim30\%$ improvement on motion forecasting of bikes, as compared to the best BEV-only \textit{MultiXNet}. Next, we compare our method to another recent multi-view method \textit{L-MV}~\cite{fadadu2020multi}. As  shown in Table~\ref{tab:nuscenes}, our method outperforms \textit{L-MV}~\cite{fadadu2020multi} on all classes by a large margin on both detection and forecasting. Building on top of \textit{MultiXNet}, \textit{L-MV} only improved the detection performance by incorporating a single sweep in RV. In contrast, we are able to utilize the temporal sequence in RV to improve both detection and motion forecasting performance. This demonstrates that our proposed method can leverage multiple views much more effectively than previous multi-view end-to-end methods. Finally, we show that our method, with only LiDAR information, is able to outperform multi-sensor methods like \textit{LiRANet}~\cite{liranet} (which uses RADAR in addition to LiDAR) and \textit{LC-MV}~\cite{fadadu2020multi} (which uses camera images in addition to LiDAR).

\begin{table*}[ht]
    \centering
    \caption{Comparison of Views in Temporal Fusion Network}
    \vspace{-0.5em}
    \scalebox{0.8}{
    \begin{tabular}{c|cc|ccc|cc|ccc|cc|ccc}
        \hline
        \multirow{3}{*}{View} & \multicolumn{5}{c|}{Vehicle} & \multicolumn{5}{c|}{Pedestrian} & \multicolumn{5}{c}{Bikes} \\ \cline{2-16}
        &  \multicolumn{2}{c|} {AP (\%) $\uparrow$} & \multicolumn{3}{c|}{$L_2$ (cm) $\downarrow$}  & \multicolumn{2}{c|} {AP (\%) $\uparrow$} & \multicolumn{3}{c|}{$L_2$ (cm) $\downarrow$}  & \multicolumn{2}{c|} {AP (\%) $\uparrow$} & \multicolumn{3}{c}{$L_2$ (cm) $\downarrow$} \\
        &  0.5 IoU & 0.7 IoU & 0.0 s & 1.0 s & 3.0 s & 0.1 IoU & 0.3 IoU & 0.0 s & 1.0 s & 3.0 s & 0.1 IoU & 0.3 IoU & 0.0 s & 1.0 s & 3.0 s \\
        \hline
        \multicolumn{16}{c}{nuScenes} \\
        \hline
        RV & 80.3 & 61.8 & 46.5 & 87.4 & 193.2 & 64.8 & 63.1 & 17.5 & 93.9 & 273.2 & 36.2 & 31.8 & 32.5 & 103.5 & 244.6 \\
        BEV & 83.2 & 65.1 & 41.5 & 57.5 & 122.4 & 70.8 & 69.0 & 16.6 & 33.7 & 84.8 & 42.5 & 37.8 & 31.1 & 58.7 & 140.9 \\
        Both & \textbf{85.1} & \textbf{67.2} & \textbf{38.8} & \textbf{53.7} & \textbf{115.9} & \textbf{73.5} & \textbf{71.9} & \textbf{16.2} & \textbf{33.2} & \textbf{84.4} & \textbf{48.0} & \textbf{43.1} & \textbf{28.7} & \textbf{52.6} & \textbf{125.1} \\
        \hline
        \multicolumn{16}{c}{Internal data set} \\
        \hline
        RV & 85.2 & 70.0 & 34.2 & 44.2 & 73.4 & 65.4 & 67.3 & 18.5 & 46.6 & 121.3 & 48.9 & 42.8 & 26.8 & 53.2 & 107.0 \\
        BEV & 88.3 & 75.0 & 29.6 & 34.4 & 55.9 & 71.8 & 69.9 & 17.6 & 31.6 & 76.4 & 48.3 & 42.6 & 26.1 & 33.3 & 56.1 \\
        Both & \textbf{89.6} & \textbf{76.7} & \textbf{27.8} & \textbf{32.4} & \textbf{53.4} & \textbf{75.6} & \textbf{73.7} & \textbf{16.9} & \textbf{30.0} & \textbf{73.4} & \textbf{57.9} & \textbf{51.4} & \textbf{24.5} & \textbf{31.7} & \textbf{54.0} \\
        \hline
    \end{tabular}
    }
    \label{tab:view_tf}
    \vspace{-0.5em}
\end{table*}

\begin{table*}[ht]
    \centering
    \caption{Comparison of Views in the Backbone Network}
    \vspace{-0.5em}
    \scalebox{0.8}{
    \begin{tabular}{c|cc|ccc|cc|ccc|cc|ccc}
        \hline
        \multirow{3}{*}{View} & \multicolumn{5}{c|}{Vehicle} & \multicolumn{5}{c|}{Pedestrian} & \multicolumn{5}{c}{Bikes} \\ \cline{2-16}
        &  \multicolumn{2}{c|} {AP (\%) $\uparrow$} & \multicolumn{3}{c|}{$L_2$ (cm) $\downarrow$}  & \multicolumn{2}{c|} {AP (\%) $\uparrow$} & \multicolumn{3}{c|}{$L_2$ (cm) $\downarrow$}  & \multicolumn{2}{c|} {AP (\%) $\uparrow$} & \multicolumn{3}{c}{$L_2$ (cm) $\downarrow$} \\
        &  0.5 IoU & 0.7 IoU & 0.0 s & 1.0 s & 3.0 s & 0.1 IoU & 0.3 IoU & 0.0 s & 1.0 s & 3.0 s & 0.1 IoU & 0.3 IoU & 0.0 s & 1.0 s & 3.0 s \\
        \hline
        \multicolumn{16}{c}{nuScenes} \\
        \hline
        RV & 84.8 & 66.67 & 39.7 & 55.12 & 120.0 & 76.1 & 74.4 & \textbf{15.6} & \textbf{31.3} & 80.3 & \textbf{50.9} & \textbf{47.2} & \textbf{27.4} & \textbf{51.8} & 128.3\\
        BEV & 85.1 & 67.2 & 38.8 & 53.7 & 115.9 & 73.5 & 71.9 & 16.2 & 33.2 & 84.4 & 48.0 & 43.1 & 28.7 & 52.6 & \textbf{125.1} \\
        Both & \textbf{85.5} & \textbf{67.8} & \textbf{38.2} & \textbf{53.1} & \textbf{115.0} & \textbf{76.4} & \textbf{74.6} & 15.9 & 31.6 & \textbf{79.9} & 49.5 & 44.5 & 28.9 & 54.3 & 131.6 \\
        \hline
        \multicolumn{16}{c}{Internal data set} \\
        \hline
        RV & 90.2 & 77.4 & 27.0 & 31.8 & 53.3 & 79.1 & 77.1 & 16.3 & 29.6 & 73.7 & 63.9 & 56.4 & 23.2 & 32.9 & 62.9 \\
        BEV & 89.6 & 76.7 & 27.8 & 32.4 & 53.4 & 75.6 & 73.7 & 16.9 & 30.0 & 73.4 & 57.9 & 51.4 & 24.5 & 31.7 & 54.0 \\
        Both & \textbf{90.8} & \textbf{78.4} & \textbf{26.1} & \textbf{30.6} & \textbf{51.4} & \textbf{79.7} & \textbf{77.8} & \textbf{16.1} & \textbf{28.8} & \textbf{71.6} & \textbf{64.5} & \textbf{57.9} & \textbf{22.7} & \textbf{30.2} & \textbf{53.2} \\
        \hline
    \end{tabular}
    }
    \label{tab:view_backbone}
    \vspace{-0.5em}
\end{table*}

\begin{table*}[!ht]
    \centering
    \caption{Comparison of Multi-View Temporal Fusion Strategies}
    \vspace{-0.5em}
    \scalebox{0.8}{
    \begin{tabular}{c|cc|ccc|cc|ccc|cc|ccc}
        \hline
        \multirow{3}{*}{Strategy} & \multicolumn{5}{c|}{Vehicle} & \multicolumn{5}{c|}{Pedestrian} & \multicolumn{5}{c}{Bikes} \\ \cline{2-16}
         &  \multicolumn{2}{c|} {AP (\%) $\uparrow$} & \multicolumn{3}{c|}{$L_2$ (cm) $\downarrow$}  & \multicolumn{2}{c|} {AP (\%) $\uparrow$} & \multicolumn{3}{c|}{$L_2$ (cm) $\downarrow$}  & \multicolumn{2}{c|} {AP (\%) $\uparrow$} & \multicolumn{3}{c}{$L_2$ (cm) $\downarrow$} \\
        &  0.5 IoU & 0.7 IoU & 0.0 s & 1.0 s & 3.0 s & 0.1 IoU & 0.3 IoU & 0.0 s & 1.0 s & 3.0 s & 0.1 IoU & 0.3 IoU & 0.0 s & 1.0 s & 3.0 s \\
        \hline
        \multicolumn{16}{c}{nuScenes} \\
        \hline
        One Shot & 84.3 & 66.3 & 40.1 & 56.0 & 120.3 & 74.5 & 72.7 & 16.1 & 33.6 & 86.1 & 46.6 & 42.2 & 29.3 & 58.4 & 142.4 \\

        Sequential & \textbf{85.5} & \textbf{67.8} & \textbf{38.2} & \textbf{53.1} & \textbf{115.0} & \textbf{76.4} & \textbf{74.6} & \textbf{15.9} & \textbf{31.6} & \textbf{79.9} & \textbf{49.5} & \textbf{44.5} & \textbf{28.9} & \textbf{54.3} & \textbf{131.6} \\
        \hline
        \multicolumn{16}{c}{Internal data set} \\
        \hline
        One Shot & 90.6  & 78.1 & 26.5 & 31.4 & 52.2 & 78.8 & 76.9 & 16.2 & 29.7 & 74.0 & 62.5 & 56.2 & 22.8 & 33.6 & 63.0 \\
        Sequential & \textbf{90.8} & \textbf{78.4} & \textbf{26.2} & \textbf{30.6} &\textbf{ 51.4} & \textbf{79.7} & \textbf{77.8} & \textbf{16.1} & \textbf{28.8} & \textbf{71.6} & \textbf{64.5} & \textbf{57.9} & \textbf{22.7} & \textbf{30.2} & \textbf{53.2} \\
        \hline
    \end{tabular}
    }
    \label{tab:temporal_fusion}
    \vspace{-0.5em}
\end{table*}

\subsection{Ablation Studies}
In this section, we analyze the impact of individual components of our multi-view fusion model. We study the individual effect of using RV and BEV information in the temporal fusion network, as well as in the backbone network. Further, we study the efficacy of our sequential fusion method for fusing multiple time-step information.

\subsubsection{Views in Temporal Fusion Network}
\vspace{-0.5em}
\label{sec:temporal_ablation}
First, we study the use of multiple views in temporal fusion, as compared to only using a single view. The RV-only baseline is created by removing the \textit{BEV Temporal Fusion} block in Figure~\ref{fig:overview-tf}. Similarly the BEV-only baseline is created by removing the \textit{RV Temporal Fusion} block in Figure~\ref{fig:overview-tf} and directly warping the input RV features into BEV without any temporal fusion in RV. For a fair comparison, we keep the same number of parameters between the single-view and multi-view experiments by moving the additional convolutions from one view to another. The results of the  comparison are shown in Table~\ref{tab:view_tf}. We observe that BEV-only fusion significantly outperforms the RV-only fusion in the task of motion forecasting. We believe this is due to the strong prior that the BEV representation provides to motion forecasting. However, after combining both views we get a large performance improvement over the BEV-only fusion model. This suggests that there is relevant information to tasks of detection and motion forecasting that is unique to each view. We also observe that the relative performance improvement on smaller objects, such as bikes ($20\%$) and pedestrians ($6\%$), is larger than on bigger objects such as vehicles ($2\%$), suggesting that the network is able to utilize the higher resolution information present in RV .

\subsubsection{Views in Backbone Network}
\vspace{-0.5em}
Next, we analyze the impact of including multiple views in the backbone network. We perform this ablation with the best performing multi-view, temporal fusion model from Section \ref{sec:temporal_ablation}. The RV-only and BEV-only baselines are created by removing the \textit{BEV Network} and \textit{RV Network} respectively in Figure~\ref{fig:overview-bb}. For the RV-only baseline we extend the detector to include two extra convolutional layers to aggregate some BEV context. Similar to the previous study, we keep the same number of parameters between single-view and multi-view experiments by moving the convolutions from one view to another. As we can see from the results in Table~\ref{tab:view_backbone}, using both RV and BEV in the backbone improves performance over any single-view method on both data sets. We further observe that the relative improvement on our internal data set is larger than on nuScenes. We believe this can be attributed to the better utilization of the $2$x higher resolution LiDAR in the RV.

\subsubsection{Strategies for Multi-View Temporal Fusion}
\vspace{-0.5em}
Finally, we compare our proposed \textit{sequential} fusion approach to the naive \textit{one-shot} approach. In contrast to \textit{sequential} warping, the \textit{one-shot} approach warps the raw per-point features from all the past sweeps directly into the RV and BEV of the reference pose. The temporal aggregation in each view is then performed by concatenating the features from all the warped views which are then used to learn independent spatio-temporal BEV and RV features and are then fed as input to the backbone network (Figure~\ref{fig:overview-bb}). The major difference lies in the absence of the sequential fusion in both views. For a fair comparison, we ensure that the number of parameters in each per-view network is same as the total parameters in the corresponding \textit{sequential} temporal fusion. 
As we can see from Table~\ref{tab:temporal_fusion}, our \textit{sequential} approach can better utilize multiple views for fusion of the temporal sequence of LiDAR data. We note that our approach has a larger relative improvement on nuScenes as compared to the internal data set. We attribute this to the fact that in nuScenes the information loss resulting from the temporal fusion stage in RV \cite{rvfusenet} has higher impact than when using higher resolution LiDAR which provides more redundancy.

\subsection{Run-time Analysis}
\vspace{-0.5em}
We report the run-time results using a Titan RTX GPU. Our method can process the operating range of $50$m on nuScenes in $\sim30$ms and the range of $100$m on our internal data set in $\sim55$ms. In contrast, the previous BEV-only method \cite{multinet} runs on the shorter range of $50$m in $\sim38$ms \cite{liranet}. BEV-only methods do not scale well with range and have not reported numbers on larger operating ranges of $100$m.

RV-only methods~\cite{rvfusenet,laserflow} have shown the ability to scale better with larger ranges than BEV-only methods. These methods reportedly process the range of $100$m in $\sim60$ms. As compared to them, we achieve faster runtime of $\sim55$ms. Therefore, our method combines the runtime advantages that RV-only methods enjoy, with better detection and motion forecasting performance of BEV-only methods. Finally, our method can finish processing all the data and produce output for each sweep before the arrival of the next sweep at 10Hz. Therefore, it is suitable for real-time on-board operations as it exhibits no latency related loss of data.

\begin{figure*}[t]
    \centering
    \includegraphics[width=0.85\textwidth]{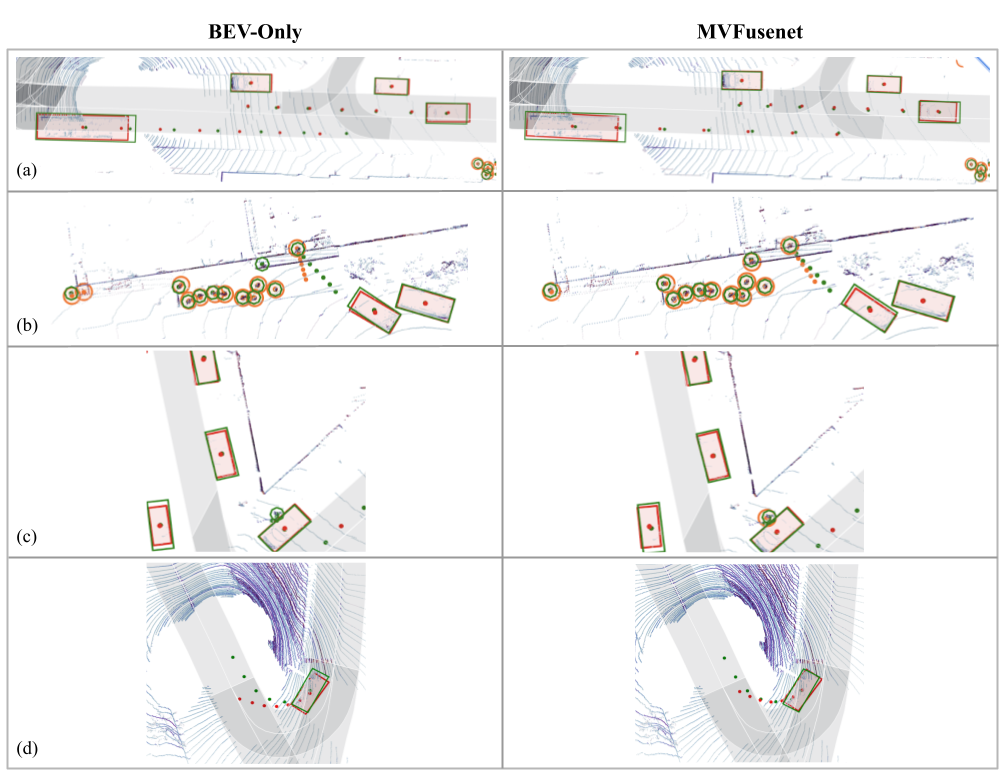}
\caption{Qualitative comparison of proposed \textit{MVFusenet} with the \textit{BEV-only} model which uses BEV in both temporal fusion and the backbone. Model outputs for detections and trajectories are depicted in \textcolor{red}{red} for vehicles and in \textcolor{orange}{orange} for pedestrians. The ground truth is depicted in \textcolor{OliveGreen}{green}. In (a), the \textit{MVFusenet} model produces better quality motion forecasts as compared to the BEV-only method for a moving vehicle (middle left). In (b), the BEV-only method exhibits multiple failure modes for pedestrians pertaining to a false positive (left), a false negative (top middle), and an inaccurate trajectory for the moving pedestrian (top right), while \textit{MVFusenet} exhibits the correct behavior. The example in (c) shows the BEV-only model failing to detect a pedestrian adjacent to the vehicle. Finally, in (d) we see that both models fail to accurately predict the position at $3$s for a vehicle turning sharply, but the proposed model more accurately predicts the turning behaviour.}

\label{fig:qualitative}
\vspace{-1em}
\end{figure*}
\subsection{Qualitative Analysis}
\vspace{-0.5em}
We present a qualitative comparison of our proposed multi-view model with a single-view BEV-only method, in Figure \ref{fig:qualitative}. While detection of vehicles is similar between the two methods, \textit{MVFusenet} more accurately detects pedestrians. Also, we show a few cases where our method is able to improve the motion prediction of vehicles and pedestrians over single-view BEV-only method. 

\section{Conclusion and Future Work}
\vspace{-0.5em}
We presented a novel multi-view model for end-to-end object detection and motion forecasting. We introduced a unique method for multi-view temporal fusion, as well as a novel multi-view backbone network. We proved the effectiveness of our approach as compared to existing single-view and  multi-view fusion methods on two large-scale data sets. We showed that the proposed method can leverage the complementary information in the RV and BEV and improve accuracy on both detection and motion forecasting tasks, while maintaining low latency and scaling to larger operating ranges. In particular, we demonstrated that incorporating both views in temporal fusion and using a sequential fusion approach significantly improves performance over only using BEV. Finally, we established a new state-of-the-art result on the publicly available nuScenes data set for joint detection and forecasting.

In addition, we have demonstrated that the presented LiDAR-only approach outperforms multi-sensor approaches which rely on RADAR or camera. However, as future work, we plan to incorporate these additional sensors to improve the robustness of the proposed approach. Additionally, we used a simple uncertainty representation and forecasting method to simplify the experimentation. In the future, we plan to incorporate recent advances in multi-modal motion forecasting and actor-scene interactions. 

\clearpage

{\small
\bibliographystyle{unsrt}
\bibliography{egbib}

\begin{thebibliography}{10}

\bibitem{spagnn}
Sergio Casas, Cole Gulino, Renjie Liao, and Raquel Urtasun.
\newblock Spatially-aware graph neural networks for relational behavior
  forecasting from sensor data.
\newblock {\em arXiv preprint arXiv:1910.08233}, 2019.

\bibitem{intentnet}
Sergio Casas, Wenjie Luo, and Raquel Urtasun.
\newblock {IntentNet}: Learning to predict intention from raw sensor data.
\newblock In {\em Proceedings of the Conference on Robot Learning (CoRL)},
  2018.

\bibitem{rvfusenet}
Ankit Laddha, Shivam Gautam, Gregory~P Meyer, and Carlos Vallespi-Gonzalez.
\newblock Rv-fusenet: Range view based fusion of time-series lidar data for
  joint 3d object detection and motion forecasting.
\newblock {\em arXiv preprint arXiv:2005.10863}, 2020.

\bibitem{multinet}
Nemanja Djuric, Henggang Cui, Zhaoen Su, Shangxuan Wu, Huahua Wang, Fang-Chieh
  Chou, Luisa~San Martin, Song Feng, Rui Hu, Yang Xu, Alyssa Dayan, Sidney
  Zhang, Brian~C. Becker, Gregory~P. Meyer, Carlos V-Gonzalez, and Carl~K.
  Wellington.
\newblock Multixnet: Multiclass multistage multimodal motion prediction, 2020.

\bibitem{liranet}
Meet Shah, Zhiling Huang, Ankit Laddha, Matthew Langford, Blake Barber, Sidney
  Zhang, Carlos Vallespi-Gonzalez, and Raquel Urtasun.
\newblock Liranet: End-to-end trajectory prediction using spatio-temporal radar
  fusion.
\newblock {\em arXiv preprint arXiv:2010.00731}, 2020.

\bibitem{laserflow}
Gregory~P Meyer, Jake Charland, Shreyash Pandey, Ankit Laddha, Carlos
  Vallespi-Gonzalez, and Carl~K Wellington.
\newblock Laserflow: Efficient and probabilistic object detection and motion
  forecasting.
\newblock {\em arXiv preprint arXiv:2003.05982}, 2020.

\bibitem{lasernet}
Gregory~P. Meyer, Ankit Laddha, Eric Kee, Carlos Vallespi-Gonzalez, and Carl~K.
  Wellington.
\newblock {LaserNet}: An efficient probabilistic 3{D} object detector for
  autonomous driving.
\newblock In {\em Proceedings of the IEEE CVPR}, 2019.

\bibitem{pointpillars}
Alex~H. Lang, Sourabh Vora, Holger Caesar, Lubing Zhou, Jiong Yang, and Oscar
  Beijbom.
\newblock {PointPillars}: Fast encoders for object detection from point clouds.
\newblock In {\em Proceedings of the IEEE CVPR}, 2010.

\bibitem{multiview}
Yin Zhou, Pei Sun, Yu~Zhang, Dragomir Anguelov, Jiyang Gao, Tom Ouyang, James
  Guo, Jiquan Ngiam, and Vijay Vasudevan.
\newblock End-to-end multi-view fusion for 3d object detection in lidar point
  clouds.
\newblock In {\em Conference on Robot Learning}, 2020.

\bibitem{fadadu2020multi}
Sudeep Fadadu, Shreyash Pandey, Darshan Hegde, Yi~Shi, Fang-Chieh Chou, Nemanja
  Djuric, and Carlos Vallespi-Gonzalez.
\newblock Multi-view fusion of sensor data for improved perception and
  prediction in autonomous driving.
\newblock {\em arXiv preprint arXiv:2008.11901}, 2020.

\bibitem{pointrcnn}
Shaoshuai Shi, Xiaogang Wang, and Hongsheng Li.
\newblock {PointRCNN}: 3{D} object proposal generation and detection from point
  cloud.
\newblock In {\em Proceedings of the IEEE CVPR}, 2019.

\bibitem{std}
Zetong Yang, Yanan Sun, Shu Liu, Xiaoyong Shen, and Jiaya Jia.
\newblock {STD}: Sparse-to-dense 3{D} object detector for point cloud.
\newblock In {\em Proceedings of the IEEE ICCV}, 2019.

\bibitem{zhouVoxelNetEndtoEndLearning2018}
Yin Zhou and Oncel Tuzel.
\newblock {VoxelNet}: End-to-end learning for point cloud based 3{D} object
  detection.
\newblock In {\em Proceedings of the IEEE CVPR}, 2018.

\bibitem{second}
Yan Yan, Yuxing Mao, and Bo~Li.
\newblock {SECOND}: Sparsely embedded convolutional detection.
\newblock {\em Sensors}, 2018.

\bibitem{pixor}
Bin Yang, Wenjie Luo, and Raquel Urtasun.
\newblock {PIXOR}: Real-time 3{D} object detection from point clouds.
\newblock In {\em Proceedings of the IEEE CVPR}, 2018.

\bibitem{zhang2020polarnet}
Yang Zhang, Zixiang Zhou, Philip David, Xiangyu Yue, Zerong Xi, Boqing Gong,
  and Hassan Foroosh.
\newblock Polarnet: An improved grid representation for online lidar point
  clouds semantic segmentation.
\newblock In {\em Proceedings of the IEEE/CVF Conference on Computer Vision and
  Pattern Recognition}, pages 9601--9610, 2020.

\bibitem{li2016vehicle}
Bo~Li, Tianlei Zhang, and Tian Xia.
\newblock Vehicle detection from 3d lidar using fully convolutional network.
\newblock {\em arXiv preprint arXiv:1608.07916}, 2016.

\bibitem{milioto2019rangenet++}
Andres Milioto, Ignacio Vizzo, Jens Behley, and Cyrill Stachniss.
\newblock Rangenet++: Fast and accurate lidar semantic segmentation.
\newblock In {\em 2019 IEEE/RSJ International Conference on Intelligent Robots
  and Systems (IROS)}, pages 4213--4220. IEEE, 2019.

\bibitem{kochanov2020kprnet}
Deyvid Kochanov, Fatemeh~Karimi Nejadasl, and Olaf Booij.
\newblock Kprnet: Improving projection-based lidar semantic segmentation.
\newblock {\em arXiv preprint arXiv:2007.12668}, 2020.

\bibitem{mv3d}
Xiaozhi Chen, Huimin Ma, Ji~Wan, Bo~Li, and Tian Xia.
\newblock Multi-view 3{D} object detection network for autonomous driving.
\newblock In {\em Proceedings of the IEEE CVPR}, 2017.

\bibitem{zhou2020end}
Yin Zhou, Pei Sun, Yu~Zhang, Dragomir Anguelov, Jiyang Gao, Tom Ouyang, James
  Guo, Jiquan Ngiam, and Vijay Vasudevan.
\newblock End-to-end multi-view fusion for 3d object detection in lidar point
  clouds.
\newblock In {\em Conference on Robot Learning}, pages 923--932. PMLR, 2020.

\bibitem{chen2020mvlidarnet}
Ke~Chen, Ryan Oldja, Nikolai Smolyanskiy, Stan Birchfield, Alexander Popov,
  David Wehr, Ibrahim Eden, and Joachim Pehserl.
\newblock Mvlidarnet: Real-time multi-class scene understanding for autonomous
  driving using multiple views.
\newblock {\em arXiv preprint arXiv:2006.05518}, 2020.

\bibitem{liong2020amvnet}
Venice~Erin Liong, Thi Ngoc~Tho Nguyen, Sergi Widjaja, Dhananjai Sharma, and
  Zhuang~Jie Chong.
\newblock Amvnet: Assertion-based multi-view fusion network for lidar semantic
  segmentation.
\newblock {\em arXiv preprint arXiv:2012.04934}, 2020.

\bibitem{liang2020rangercnn}
Zhidong Liang, Ming Zhang, Zehan Zhang, Xian Zhao, and Shiliang Pu.
\newblock Rangercnn: Towards fast and accurate 3d object detection with range
  image representation.
\newblock {\em arXiv preprint arXiv:2009.00206}, 2020.

\bibitem{dp}
Nemanja Djuric, Vladan Radosavljevic, Henggang Cui, Thi Nguyen, Fang-Chieh
  Chou, Tsung-Han Lin, and Jeff Schneider.
\newblock Motion prediction of traffic actors for autonomous driving using deep
  convolutional networks.
\newblock {\em arXiv preprint arXiv:1808.05819}, 2018.

\bibitem{sociallstm}
Alexandre Alahi, Kratarth Goel, Vignesh Ramanathan, Alexandre Robicquet,
  Li~Fei-Fei, and Silvio Savarese.
\newblock {Social LSTM}: Human trajectory prediction in crowded spaces.
\newblock In {\em Proceedings of the IEEE CVPR}, 2016.

\bibitem{convolutionalsocialpooling}
Nachiket Deo and Mohan~M Trivedi.
\newblock Convolutional social pooling for vehicle trajectory prediction.
\newblock In {\em Proceedings of the IEEE CVPR Workshops (CVPRW)}, 2018.

\bibitem{desire}
Namhoon Lee, Wongun Choi, Paul Vernaza, Christopher~B Choy, Philip~HS Torr, and
  Manmohan Chandraker.
\newblock {DESIRE}: Distant future prediction in dynamic scenes with
  interacting agents.
\newblock In {\em Proceedings of the IEEE CVPR}, 2017.

\bibitem{lasernet++}
Gregory~P. Meyer, Jake Charland, Darshan Hegde, Ankit Laddha, and Carlos
  Vallespi-Gonzalez.
\newblock Sensor fusion for joint 3{D} object detection and semantic
  segmentation.
\newblock In {\em Proceedings of the IEEE CVPR Workshops (CVPRW)}, 2019.

\bibitem{destination}
Karttikeya Mangalam, Harshayu Girase, Shreyas Agarwal, Kuan-Hui Lee, Ehsan
  Adeli, Jitendra Malik, and Adrien Gaidon.
\newblock It is not the journey but the destination: Endpoint conditioned
  trajectory prediction.
\newblock {\em arXiv preprint arXiv:2004.02025}, 2020.

\bibitem{zhao2020tnt}
Hang Zhao, Jiyang Gao, Tian Lan, Chen Sun, Benjamin Sapp, Balakrishnan
  Varadarajan, Yue Shen, Yi~Shen, Yuning Chai, Cordelia Schmid, et~al.
\newblock Tnt: Target-driven trajectory prediction.
\newblock {\em arXiv preprint arXiv:2008.08294}, 2020.

\bibitem{multipath}
Yuning Chai, Benjamin Sapp, Mayank Bansal, and Dragomir Anguelov.
\newblock {MultiPath}: Multiple probabilistic anchor trajectory hypotheses for
  behavior prediction.
\newblock {\em arXiv preprint arXiv:1910.05449}, 2019.

\bibitem{covernet}
Tung Phan-Minh, Elena~Corina Grigore, Freddy~A Boulton, Oscar Beijbom, and
  Eric~M Wolff.
\newblock Covernet: Multimodal behavior prediction using trajectory sets.
\newblock In {\em Proceedings of the IEEE CVPR}, 2020.

\bibitem{rhinehart2019precog}
Nicholas Rhinehart, Rowan McAllister, Kris Kitani, and Sergey Levine.
\newblock Precog: Prediction conditioned on goals in visual multi-agent
  settings.
\newblock In {\em Proceedings of the IEEE/CVF International Conference on
  Computer Vision}, pages 2821--2830, 2019.

\bibitem{mfp}
Charlie Tang and Russ~R Salakhutdinov.
\newblock Multiple futures prediction.
\newblock In {\em Proceedings of Advances in Neural Information Processing
  Systems (NIPS)}, 2019.

\bibitem{cui2019multimodal}
Henggang Cui, Vladan Radosavljevic, Fang-Chieh Chou, Tsung-Han Lin, Thi Nguyen,
  Tzu-Kuo Huang, Jeff Schneider, and Nemanja Djuric.
\newblock Multimodal trajectory predictions for autonomous driving using deep
  convolutional networks.
\newblock In {\em 2019 International Conference on Robotics and Automation
  (ICRA)}. IEEE, 2019.

\bibitem{gao2020vectornet}
Jiyang Gao, Chen Sun, Hang Zhao, Yi~Shen, Dragomir Anguelov, Congcong Li, and
  Cordelia Schmid.
\newblock Vectornet: Encoding hd maps and agent dynamics from vectorized
  representation.
\newblock In {\em Proceedings of the IEEE/CVF Conference on Computer Vision and
  Pattern Recognition}, pages 11525--11533, 2020.

\bibitem{yu2020spatio}
Cunjun Yu, Xiao Ma, Jiawei Ren, Haiyu Zhao, and Shuai Yi.
\newblock Spatio-temporal graph transformer networks for pedestrian trajectory
  prediction.
\newblock In {\em European Conference on Computer Vision}, pages 507--523.
  Springer, 2020.

\bibitem{gupta2018social}
Agrim Gupta, Justin Johnson, Li~Fei-Fei, Silvio Savarese, and Alexandre Alahi.
\newblock Social gan: Socially acceptable trajectories with generative
  adversarial networks.
\newblock In {\em Proceedings of the IEEE Conference on Computer Vision and
  Pattern Recognition}, pages 2255--2264, 2018.

\bibitem{lanegcn}
Ming Liang, Bin Yang, Rui Hu, Yun Chen, Renjie Liao, Song Feng, and Raquel
  Urtasun.
\newblock Learning lane graph representations for motion forecasting.
\newblock {\em arXiv preprint arXiv:2007.13732}, 2020.

\bibitem{faf}
Wenjie Luo, Bin Yang, and Raquel Urtasun.
\newblock Fast and furious: Real time end-to-end {3D} detection, tracking and
  motion forecasting with a single convolutional net.
\newblock In {\em Proceedings of the IEEE CVPR}, 2018.

\bibitem{resnet}
Kaiming He, Xiangyu Zhang, Shaoqing Ren, and Jian Sun.
\newblock Deep residual learning for image recognition.
\newblock In {\em Proceedings of the IEEE CVPR}, 2016.

\bibitem{centerdet}
Tianwei Yin, Xingyi Zhou, and Philipp Kr{\"a}henb{\"u}hl.
\newblock Center-based 3d object detection and tracking.
\newblock {\em arXiv preprint arXiv:2006.11275}, 2020.

\bibitem{retinanet}
Tsung-Yi Lin, Priya Goyal, Ross Girshick, Kaiming He, and Piotr Doll{\'a}r.
\newblock Focal loss for dense object detection.
\newblock In {\em Proceedings of the IEEE ICCV}, 2017.

\bibitem{lasernet-kl}
Gregory~P Meyer and Niranjan Thakurdesai.
\newblock Learning an uncertainty-aware object detector for autonomous driving.
\newblock {\em arXiv preprint arXiv:1910.11375}, 2019.

\bibitem{nuscenes}
Holger Caesar, Varun Bankiti, Alex~H Lang, Sourabh Vora, Venice~Erin Liong,
  Qiang Xu, Anush Krishnan, Yu~Pan, Giancarlo Baldan, and Oscar Beijbom.
\newblock {NuScenes}: A multimodal dataset for autonomous driving.
\newblock {\em arXiv preprint arXiv:1903.11027}, 2019.

\bibitem{paszke2019pytorch}
Adam Paszke, Sam Gross, Francisco Massa, Adam Lerer, James Bradbury, Gregory
  Chanan, Trevor Killeen, Zeming Lin, Natalia Gimelshein, Luca Antiga, et~al.
\newblock Pytorch: An imperative style, high-performance deep learning library.
\newblock {\em arXiv preprint arXiv:1912.01703}, 2019.

\end{thebibliography}
}

\end{document}